\documentclass[sigconf]{acmart}
\AtBeginDocument{%
  \providecommand\BibTeX{{%
    \normalfont B\kern-0.5em{\scshape i\kern-0.25em b}\kern-0.8em\TeX}}
}

\usepackage{pgfplots}
\usepackage{comment}
\usepackage{hyperref}
\usepackage{subcaption}
\usepackage{booktabs} 
\usepackage{commath}
\usepackage{amsmath}
\usepackage{mathtools}
\usepackage{multirow}
\usepackage[ruled,vlined,noresetcount]{algorithm2e}
\usepackage{colortbl}
\usepackage[inline]{enumitem}

\newcommand{\para}[1]{\paragraph{\textnormal{\textbf{#1}}}}
\renewcommand{\vec}[1]{\mathbf{#1}}

\newcommand\tabfitcolw{\resizebox{1.00\columnwidth}{!}}

\copyrightyear{2021}
\acmYear{2021}
\setcopyright{acmcopyright}\acmConference[CIKM '21]{Proceedings of the 30th ACM International Conference on Information and Knowledge Management}{November 1--5, 2021}{Virtual Event, QLD, Australia}
\acmBooktitle{Proceedings of the 30th ACM International Conference on Information and Knowledge Management (CIKM '21), November 1--5, 2021, Virtual Event, QLD, Australia}
\acmPrice{15.00}
\acmDOI{10.1145/3459637.3482146}
\acmISBN{978-1-4503-8446-9/21/11}

\begin{document}
\fancyhead[]{}
\title{Multi-Objective Few-shot Learning for Fair Classification}


\author{Ishani Mondal}\authornote{Work started during the internship of the first author at IBM Research, Dublin.}
\affiliation{
\institution{Microsoft Research Lab}
\city{Bangalore}
\country{India}
}
\email{t-imonda@microsoft.com}

\author{Procheta Sen}
\affiliation{%
\institution{Dublin City University}
\city{Dublin}
\country{Ireland}
}
\email{procheta.sen@adaptcentre.ie}

\author{Debasis Ganguly}
\affiliation{%
\institution{University of Glasgow}
\city{Glasgow}
\country{U.K.}
}
\email{Debasis.Ganguly@glasgow.ac.uk}

\begin{abstract}

In this paper, we propose a general framework for mitigating the disparities of the predicted classes with respect to secondary attributes within the data (e.g., race, gender etc.).
Our proposed method involves learning a multi-objective function that in addition to learning the primary objective of predicting the primary class labels from the data, also employs a clustering-based heuristic to minimize the disparities of the class label distribution with respect to the cluster memberships, with the assumption that each cluster should ideally map to a distinct combination of attribute values. Experiments demonstrate effective mitigation of cognitive biases on a benchmark dataset without the use of annotations of secondary at-tribute values (the zero-shot case) or with the use of a small number of attribute value annotations (the few-shot case).

\end{abstract}



\settopmatter{printfolios=true}

\maketitle
\pagenumbering{gobble}
Ethics has a a pivotal role to play in shaping up the future of AI \citep{Etzioni}. Ensuring ethical decisions is particularly challenging for deep (data-driven) models because these systems are reported to be vulnerable to cognitive biases present within the data \citep{zhao-etal-2017-men,Bolukbasi:2016}, e.g., predictions such as African-Americans are more likely to commit crimes \citep{ProPublica}, or that men possessing greater erudite skills than women \citep{Bolukbasi:2016} etc.

A debiasing approach should essentially achieve a trade-off between compromising on the accuracy of a predictive model on the one hand, and that of intentionally allowing \emph{calculated error} into the predictions to achieve a reasonable degree of fairness \citep{Zafar:2017}.
%
Existing approaches employ a multi-objective likelihood (loss) function to achieve a trade-off between the primary task effectiveness and a quantitative measure of the bias evident in the system predictions \citep{Zafar:2017,sen}. Quantification of bias usually relies on the existence of annotated data resources in the form of attribute value pairs associated with the data, e.g., authorship attribution information comprising age, gender or other information \citep{Koppel}.

\para{Our Contributions}
Since attribute-value augmented datasets are difficult to construct in practice at a large scale for a wide variety of domains due to resource constraints, a scalable solution for debiasing prediction models is to apply self-retrospection.
More concretely, for bias mitigation we define the fairness criteria for the predictive loss function with the help of a cluster-based heuristic as opposed to using annotated attribute-value pairs (e.g. as in \cite{sen}). As input, our method requires the information only on the number of different values an attribute can possibly assume (e.g., 2 for `gender', 3 for `race' etc.). In our experiments, we also show that the effectiveness of this debiasing technique can further be improved with the presence of a small seed set of of annotations for the attribute value pairs.

%



\section{Proposed Debiasing Approach} 
\subsection{Problem Definition}


In a supervised deep classification task, the input is of the form $\vec{x} \in \mathbb{R}^d$, which is an embedded (pre-trained) representation of raw data, $\vec{w}$ (e.g., words in text or pixel intensities in images),
and the output is a variable of the form $y \in \mathbb{Z}_c$ comprising $c$ integer categories or labels, the end-task being to
estimate $\phi: \vec{x} \mapsto y$.

\para{Attributes of Data Instances}
It is useful to interpret bias as a property emerging from the distribution of binary relations between the primary task labels and attributes associated with the data. To formally define the relation between the categories of a prediction task, an attribute (property) of the set of data instances, let us assume that each data instance, $\vec{x}$ is associated with a categorical vector corresponding to values of $M$ attributes, $\vec{z}=(Z_1,\ldots,Z_M) \in \mathbb{Z}^{p_1}\ldots\times \mathbb{Z}^{p_m}$.
The modulo notation indicates that the number of possible values of the $i^{th}$ attribute, $Z_i$, is $p_i$, the values themselves being (without loss of generality) $0,\ldots,p_i-1$.

\para{Biased Predictions}
To define bias for a particular class label with respect to a set of inputs, we quantify bias as a non-uniformity or imbalance in the distribution of the posteriors. Specifically,
the bias for a particular combination of class label $y=a$ and an attribute value $Z_i=b$ 
can be computed as
\begin{equation}
y^{(a, b)}(\vec{x}) =
\begin{cases}
1, & \mathbb{I}[y=a \land Z_i=b]/\mathbb{I}[y=a] \geq \epsilon \\
0, & \mathrm{otherwise}
\end{cases}
\label{eq:biaswithannot}
\end{equation}
where
$\mathbb{I}(\mathcal{X})$ denotes the count of instances with property $\mathcal{X}$, and 
$\epsilon \in [0, 1]$ is a tolerance parameter for non-uniformity in the posterior.
The posterior distribution of Equation \ref{eq:biaswithannot} requires an explicit annotation of the attribute values for each input instance, which may not be available in practice. The next section discusses ways to alleviate this limitation.

\subsection{Zero-shot Approach}  \label{ss:zeroshot}

Despite the predicted category being not an explicit function of an attribute value, the presence of a particular bias (Equation \ref{eq:biaswithannot}) in the posterior with respect to an attribute value indicates that there is a latent relationship between a part of the input data representation with the attribute value. For example, in a dataset, although the gender attribute value - `female' may not be explicitly annotated in the data, the predicted output may still be `fear' for most sentences the the presence of the female pronouns.

We hypothesize that similar data instances (specifically those with identical primary task class labels, e.g. $y=a$) are mapped to similar attribute values.
If $X^a$ denotes the set of data instances with the given primary task label $a$, then for an attribute, $Z_i \in \mathbb{Z}^{p_i}$ (comprised of $p_i$ categories), each partition $X^a_j$ of $X^a$ is assumed to be associated with one of the $p_i$ categories, i.e.,
\begin{equation}
\theta_{Z_i}: X^a_j \mapsto j,\quad \cup_{j=1}^{p_i} X^a_j = \{\forall \vec{x} \in X: y_i=a\} = X^a.
\label{eq:bias_wo_annots_twovarcase}
\end{equation}

\subsubsection{Posteriors with Cluster Memberships}
The partitioning of the set of inputs, $X^a$, into $K$ clusters provides a way of estimating the posterior of Equation \ref{eq:bias_wo_annots_twovarcase}
by substituting the counts of Equation \ref{eq:biaswithannot} with the cluster membership counts. Formally,
\begin{equation}
y^{(a, X^a_j)}(\vec{x}) =
\begin{cases}
1, & \mathrm{abs}(|X^a_j|- \lfloor|X^a|/K\rfloor)/\lfloor|X^a|/K\rfloor \geq \epsilon \\
0, & \mathrm{otherwise}.
\end{cases}
\label{eq:biaswoannot}
\end{equation}

The following points are to be noted for Equation \ref{eq:biaswoannot}. First, the bias in Equation \ref{eq:biaswoannot} is defined with respect to an output class label and a partition, instead of an attribute value, as in Equation \ref{eq:biaswithannot}.  
Second, the numerator, $\mathrm{abs}(|X^a_j|- \lfloor|X^a|/K\rfloor)$, of Equation \ref{eq:biaswoannot} counts the shift in the cardinality of a cluster from the uniform cardinality (i.e. if each cluster were to be of the same size). This when divided by the denominator of the uniform count gives a relative shift normalized in $[0, 1]$, which allows provision to apply a threshold on this value similar to Equation \ref{eq:biaswithannot}.

Further, a generalization of Equation \ref{eq:bias_wo_annots_twovarcase} to $M$ attributes would mean that 
the set $X^a$ now needs to be partitioned into a total of $K=p_1p_2\ldots p_m$ number of clusters, i.e.,
\begin{equation}
\theta_{\{Z_1,\ldots,Z_M\}}:
X^a_j \mapsto j,\quad  \cup_{j=1}^{K}X^a_j = \{
\vec{x} \in X:  y_i=a\} = X^a.
\label{eq:bias_wo_annots_mvarcase}
\end{equation}

\begin{figure}[t]
    \centering
    \includegraphics[width=.8\columnwidth]{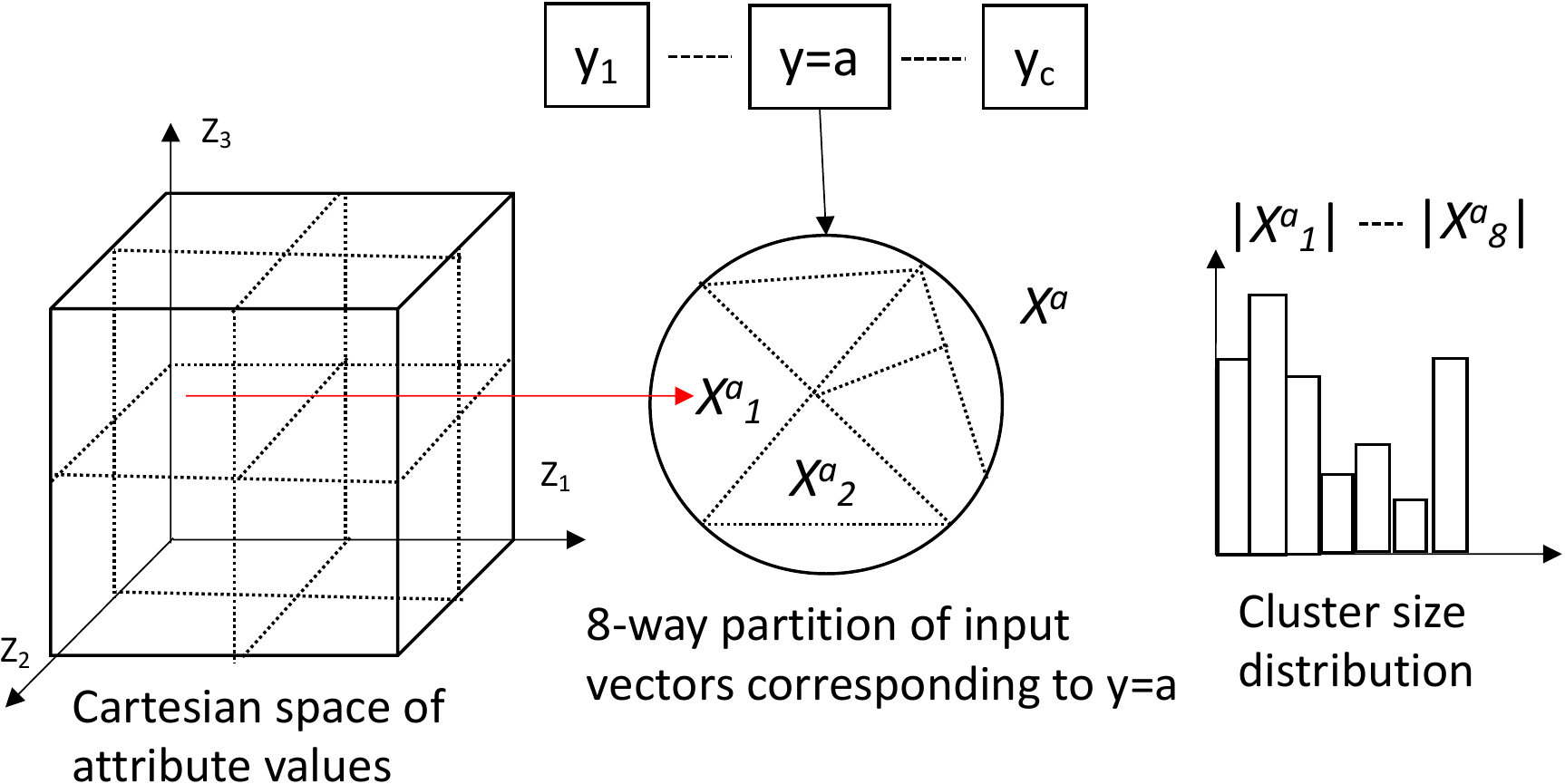}
    \caption{\small Illustration of how a particular combination of attribute values correspond to a partition of the input subset with a particular primary-task class label.}
    \label{fig:partition}
\end{figure}

The general case is schematically illustrated in Figure \ref{fig:partition}, which shows an example partition of the input of set of vectors, $X$, first into a subset for which the primary label is a particular value (say $y=a$), this subset being denoted by $X^a$.
The figure then also shows how each of these subsets, $X^a$, is
further partitioned
into a number of clusters determined by the value $K=p_1p_2p_3$.
In our example of Figure \ref{fig:partition}, this value is $2^3=8$ as shown by the cube corresponding to $Z_1\times Z_2 \times Z_3$. This pictorially depicts the hypothesis that a cube from the attribute-category space maps onto a subset, $X^a$, of the input with a particular value of the output label ($y=a$), e.g.,
the cube $(0, 0, 1)$ maps onto $X^a_1$ and so on.

\subsection{Few-Shot Approach} \label{ss:fewshot}

A point to note is that in our proposed methodology of unsupervised or zero-shot debiasing, the detection of bias in the absence of attribute value annotations depends only on the \emph{number of partitions} of $X^a$, which in turn depends on the number of category values that each attribute can assume.
The posterior probabilities themselves do not depend on which particular partition of an attribute value combination maps on to which particular partition of an output label value, e.g.,
with reference to Figure \ref{fig:partition}, these probabilities do not depend on which particular cube from the left maps onto which particular partition on the right. This limitation exists
because, in practice, such a mapping cannot be learned without the presence of any input to attribute value annotations.

To alleviate this limitation,
we now extend the unsupervised methodology of Section \ref{ss:zeroshot} to a semi-supervised one, where only a small number of data instances with additional knowledge in the form of attribute value annotations may be used to better estimate the primary label vs. attribute proportions for the purpose of mitigating biased predictions.
Given a seed set of data instances with annotated values for the secondary attributes (e.g., race, gender etc.), we first use the distinct combinations of these attribute values to construct the initial partition (e.g. instances with race=`Caucasian' and gender=`female' defines one cluster and so on). Next, for the remaining instances in the training set, we use a $K$-nearest neighbour (NN) approach to assign cluster memberships, i.e. an attribute of each non-seed instance is assigned to the majority attribute value found in its $k$-NN neighborhood (with the usual notion of cosine distance between embedded data representations). 
%

The disparity in the cluster memberships of Equation \ref{eq:bias_wo_annots_twovarcase} is then computed in the same way, the only difference being that the formation of the clusters in this case is \emph{guided by} the external knowledge of attribute values provided in the form of a seed set.

\subsection{Debiasing with Multi-task Objective}


%

After the process of clustering either
in the unsupervised or the semi-supervised way (Sections \ref{ss:zeroshot} and \ref{ss:fewshot} respectively), we use the existing multi-objective based approach of \cite{sen} for mitigating the bias in the primary classification task.
The objective of the multi-task learning is our context is to learn to correctly predict the primary task labels, and \emph{incorrectly} predict the bias indicator labels, i.e., in other words, invert the bias indicators for specific bias units (combinations of primary category label and attribute values).
%
Formally, the following loss is minimized
\begin{equation}
\mathcal{L}(\vec{x},\Theta_p,\Theta_B,\beta)= 
\sigma(\Theta_p\cdot\vec{x})_a 
- \sigma(\Theta_B\cdot\vec{x}) 
+ \beta||\Theta_p||_2, 
\label{eq:jointloss}
\end{equation}
where $\sigma(\Theta_p\cdot\vec{x})_a$ denotes the softmax probability corresponding to the label $y=a$, $\sigma(\Theta_B\cdot\vec{x})$ is the sigmoid indicating presence of bias,
and $\beta$ is a $L_2$ regularization parameter for the primary task.

\section{Evaluation}

\subsection{Experiment Setup}

Following the prescribed experiment setup on bias mitigation, we, in our experiments, also use a balanced emotion prediction dataset, named the Equity Evaluation Corpus (EEC) \citep{kiritchenko-mohammad-2018-examining}.
%
%
The primary task in the EEC dataset is to predict the emotion associated with each sentence, the four emotion categories being `joy', `fear', `sadness' and `anger'.
Using the notations of Equation \ref{eq:biaswithannot}, for this dataset $M=2$ (two attributes - $Z_1$ and $Z_2$), with $p_1=2$ (Male/Female) and $p_2=2$ (Caucasian/African-American).
%
%
A property of the dataset is that the distribution of the sentences of the EEC dataset
with respect to the emotion and the demographics attributes is almost uniform. Similar to \cite{sen},
as input representation of each sentence, we used the sum of the pre-trained embedded vectors of its constituent words (300 dimensional skip-gram vectors from Google-News).

Fairness, or lack of bias, is measured by a posterior estimate similar to that of Equation \ref{eq:biaswithannot} with two differences - i) we use predicted labels, $\hat{y}$'s, instead of the reference ones, and ii) we do not apply the threshold over the posteriors. Formally speaking, we define the metric \emph{fairness}, $F{(a,Z_i)}$, with respect to a label $\hat{y}=a$ and an attribute $Z_i$ as
\begin{equation}
F(a,Z_i)=p_i^{p_i}\prod_{b \in \mathbb{Z}_{p_i}}\frac{\mathbb{I}[\hat{y}=a \land Z_i=b]}{\mathbb{I}[\hat{y}=a]}, \label{eq:fairness}   
\end{equation}
the maximum value of which occurs when each of the $p_i$ probabilities equals $1/p_i$ (the normalization constant $p_i^{p_i}$ scaling the maximum value to $1$), i.e. when the posteriors are perfectly uniform.
%
%
%
In addition to measuring fairness, we also report accuracy, $A$, of the prediction task. We also combine the two into a single metric, $\gamma$, where $\gamma=2FA/(F+A)$. 



\subsection{Methods Investigated}

The different methods, investigated in our experiments, are loosely classified into one of the following -
\begin{enumerate*}[label=\textit{\alph*)}]
\item \emph{bias-agnostic}, methods that do not explicitly model the bias from predicted outputs,
\item \emph{zero and few-shot bias-aware}, methods, which either do not use attribute annotations or use a small set of them,
and 
\item \emph{supervised bias-aware}, supervised methods that explicitly use attribute annotations for each data instance in the training set for debiasing.
\end{enumerate*}

\para{Bias-agnostic approaches} 
The objective of using bias-agnostic approaches is to see if a general bias-agnostic technique, such as regularization, could be effective in mitigating bias from downstream tasks. In \textbf{Bias-Agnostic Classification (BAC)} we experiment with a standard $L_2$ regularization to see if intentionally introducing a small quantity of noise during training can help mitigate the bias. Specifically, this is an ablation baseline of Equation \ref{eq:jointloss}, where the $\Theta_B$ part of the network is absent. 

In \textbf{Bias-Agnostic dynamic regularization (BADR)} methods, we set a value of the regularization parameter as a function of the cluster membership based posterior distribution (Equation \ref{eq:biaswoannot}). This baseline is an ablation study for our proposed method, where instead of setting the bias indicating variables in the multi-objective loss of Equation \ref{eq:jointloss}, we use the cluster membership distribution only to relax the overfitting. Specifically, instead of using a constant value of 
$\beta$ in Equation \ref{eq:jointloss}, we set $\beta=\beta_0 \max_j \mathrm{abs}(|X^a_j|- \lfloor|X^a|/K\rfloor)/\lfloor|X^a|/K\rfloor$ (see Equation \ref{eq:biaswoannot}) (i.e., the maximum relative shift from uniformity in the cluster membership based posteriors) so as to allow higher regularization for more skewed (biased) posteriors. The value of 
$\beta_0$
in this setup is the value of 
$\beta$ which yielded the maximum $\gamma$ (harmonic mean of accuracy and fairness). 

\para{Bias-Aware Supervised Approaches}
These methods make use of the external knowledge of the attribute values and thereby serve as providing an upper bound on the effectiveness of debiasing results.
The different methods in this category are described as follows.
For \textbf{Bias-Aware Supervised (BAS)} methods, we employ the multi-tasking architecture of \cite{sen} along with the attribute annotations of gender and ethnicity categories. The method \textbf{Bias-Aware Debiased Word Embedding Approach (BAS-DW)}, is identical to that of \textbf{BAS}, the only difference being that debiased word-embedding is used as inputs.
 In fact, as per our naming convention, the suffix `DW' attached to a method's name indicates the application of the same method with debiased word embeddings provided as inputs \citep{Bolukbasi:2016}.

\para{Zero-shot and Few-shot Approaches} 
The \textbf{Bias-aware Cluster based Posterior (BACP)} is our proposed zero-shot method which does not make use of any external knowledge of attribute-values. Instead, we use the cluster membership based posteriors ($K$-means clustering of the sentence vectors grouped according to emotion labels) of Equation \ref{eq:biaswoannot} to set up the secondary pseudo-task of modeling the inverse of the bias variables as auxiliary labels (Equation \ref{eq:jointloss}).
%
For the few-shot setup, we make use of the annotations for a small proportion $\alpha \in [0, 1]$, of the training set.
In \textbf{Bias-Aware with Seed Attribute-values (BASAV)},
we make use of a semi-supervised approach to first train a gender and a race prediction classifier on the seed set of instances, following which we  infer the gender and the race attributes for each non-seed instance.
%
%
%
In \textbf{Bias-Aware KNN with Seed Attribute-values (BASAV-KNN)}, instead of applying a classifier trained on the seed set, we employ
the $k$-NN based input space partitioning of the training set to estimate the auxiliary bias labels (see Section \ref{ss:fewshot}).

\begin{table}[t]
\centering
\tabfitcolw
{
\begin{tabular}
{@{}llrccccccccrr@{}}
\toprule
& &\multicolumn{2}{c}{(Fear, Gender)}  &  \multicolumn{2}{c}{(Fear, Race)} & \multicolumn{2}{c}{(Anger, Gender)}   \\
\cmidrule(r){3-4}
\cmidrule(r){5-6}
\cmidrule(r){7-8}

Method & Acc  &  F & $\gamma$  &  F & $\gamma$ &   F & $\gamma$  \\
\midrule
BAC & 0.8238 & 0.6818 & 0.7461 & 0.8724 & 0.8474 & 0.8624 & 0.8426  \\
BADR & 0.8159 & 0.6842 & 0.7442 & 0.8724 & 0.8432 & 0.8811 & 0.8472 \\


\midrule
BAS & \textbf{0.8870} & 0.9532 & 0.9189 & 0.9560 & 0.9202 & 0.9756 & 0.9291 \\
BAS-DW &  0.8503 & 0.9474 & 0.8962 & 0.9121 & 0.8801 & 0.9703 & 0.9063\\
\midrule



BACP & 0.8524 & 0.9849 & 0.9138 & 0.9855 & 0.9141 & \textbf{0.9997} & 0.9202 \\
BACP-DW & 0.8211 & 0.9515 & 0.8815 & 0.9532 & 0.8822 & 0.9744 & 0.8912 \\


\midrule







BASAV &  0.8633 & 0.9954 & 0.9246 & 0.994 & 0.9240 &  0.999 & 0.9262 \\
BASAV-DW &  0.8189 & 0.9563 & 0.8822 & 0.985 & 0.8943 &  0.9743 & 0.8898 \\

BASAV-KNN & 0.8782 & \textbf{0.9993} & \textbf{0.9348} & \textbf{0.9913} & \textbf{0.9313} &  0.9992 & \textbf{0.9348} \\
BASAV-KNN-DW & 0.8249 & 0.9721 & 0.8924  & 0.9642 & 0.8891 & 0.9611 & 0.8878\\



\bottomrule
\end{tabular}
}
\caption{
\small
Results of bias mitigation
on the EEC dataset with various approaches.
The table reports the optimal results for each method, e.g. for the few-shot setup best results were obtained with $\alpha=0.2$.
}

\label{tab:results}
\end{table}

\subsection{Results}

Table \ref{tab:results} shows a summary of the best results obtained with each approach. We tabulate the results of fairness, accuracy and $\gamma$ for the best settings of each method. We report fairness metric for the three different combinations of emotion category and attributes that correspond to societal biases, e.g., fear:women (predictions that women are prone to be afraid) etc.

\begin{figure}[t]
\centering
\includegraphics[width=0.48\columnwidth ]{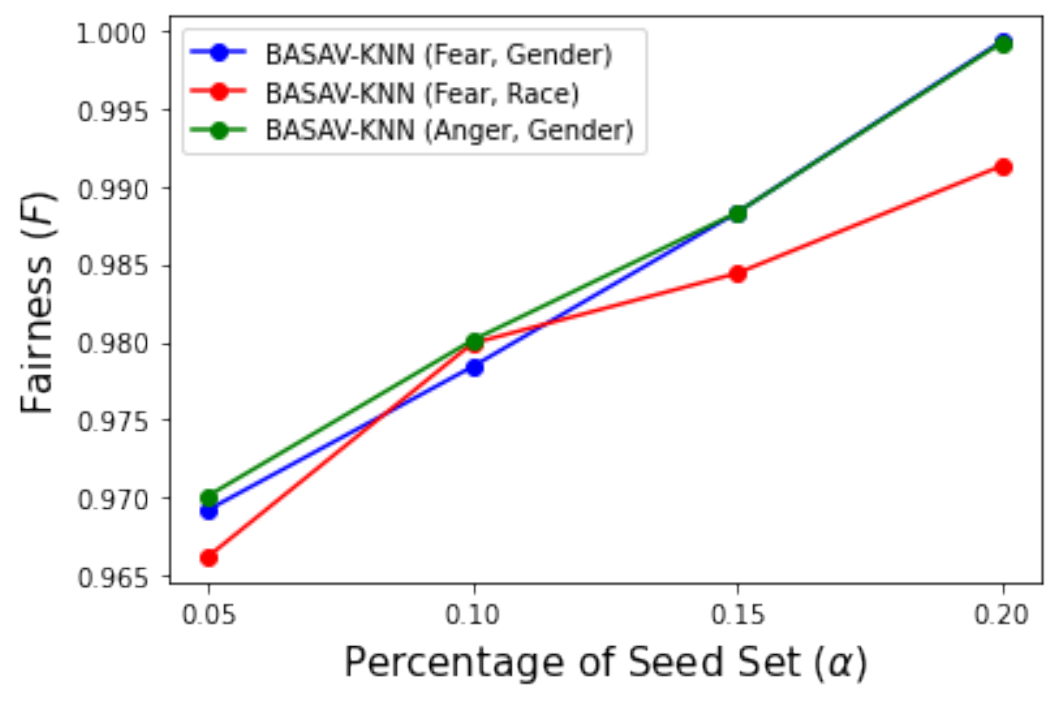}
\includegraphics[width=0.48\columnwidth]{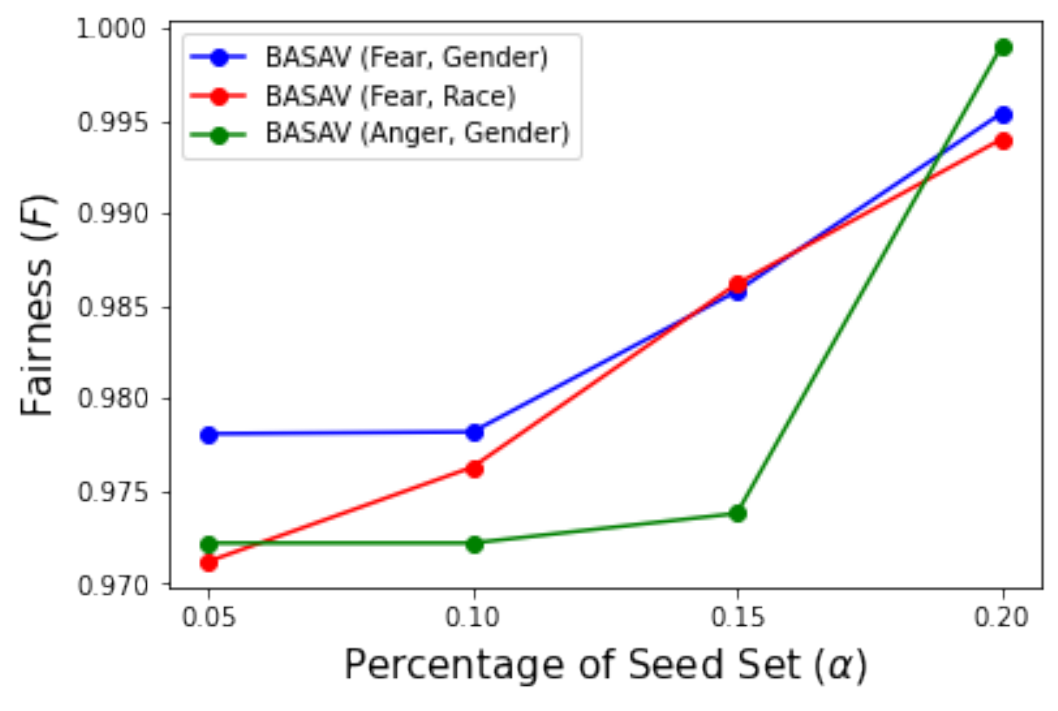}
\caption{
\small
Comparison of a parametric and a non-parametric approach for leveraging information from the seed set in a few-shot debiasing setup (respectively, left and right).
}
\label{fig:fewshot}
\end{figure}

\begin{figure}[t]
\centering
\includegraphics[width=0.48\columnwidth]{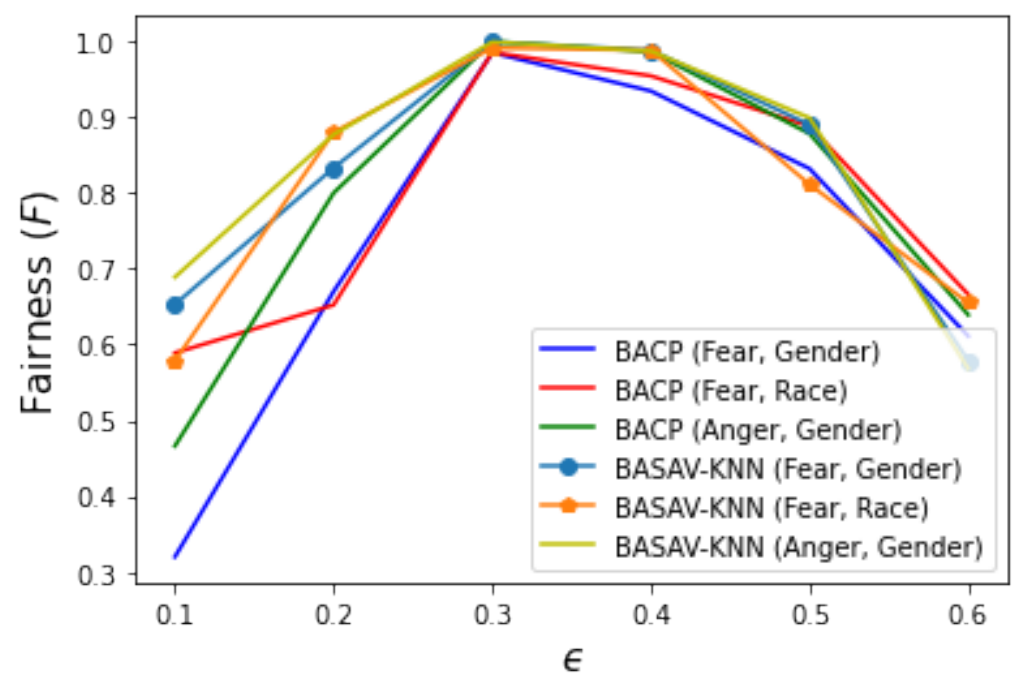}
\includegraphics[width=0.48\columnwidth]{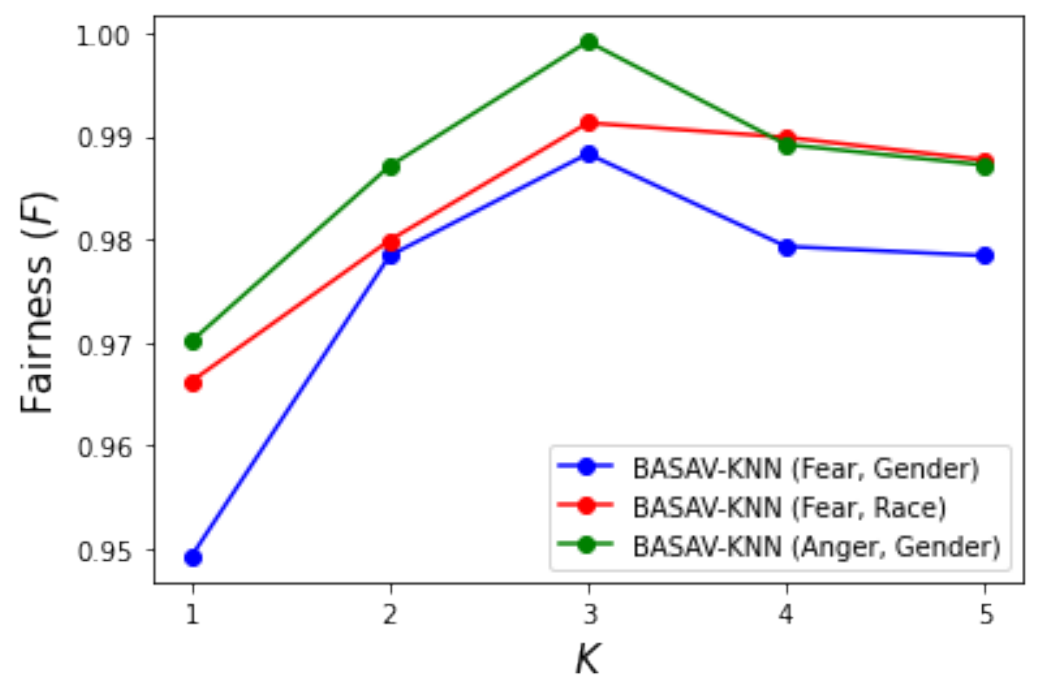}
\caption{
\small
Sensitivity of fairness with respect to variations in the values of $\epsilon$ (threshold of disparity in the posteriors shown in Equation \ref{eq:biaswoannot}) and $K$ for $K$-NN based cluster assignment, shown in the left and right plots, respectively.}
\label{fig:epsilonVary1}
\end{figure}

First, we observe that although bias-agnostic approaches generally result in effective classification,
these approaches lead to biased predictions (as seen from the low values of $F$). 
Another observation from Table \ref{tab:results} is that methods that debias the input (instead of debiasing a downstream task) do not work well in practice,
e.g. see the results where debiased word embedded vectors were used as inputs (compare BAS-DW to BAS, and BACP-DW to BACP etc.).

Supervised multi-tasking based methods (e.g., BAS etc.), that make use of the annotated attribute values during training lead to the most effective debiasing (high values of fairness and accuracy). However, these
methods leverage additional information from attribute values, which is likely not be available in a large scale classification setup.

In our proposed methods of zero-shot setup, we observe that even without making use of any annotations, the clustering-based heuristic to set the variables (Equation \ref{eq:biaswoannot}) for the pseudo-task of bias removal (Equation \ref{eq:jointloss})
achieves comparable results with the supervised approaches.
Further, we observe that the $K$-NN partitioning based semi-supervised approach achieves the best fairness (and $\gamma$) with the use of 20\% seed annotations (i.e. $\alpha=1/5$).
This shows that the clustering heuristic turns out to be more effective when the partitioning is guided by a seed set of attribute values than being purely determined by the embedded representation of the data instances themselves.
Moreover, the reason that the partitioning based method outperforms a classifier based attribute inference one (i.e., BASAV) is that the parametric models usually require a large amount of data for effective training. On the other hand, a non-parametric method such as $K$-NN is relatively insensitive to the amount of the training data (the availability of which is scarce in this setup). 
%
Similar trends are also observed for other categories of biases, such as fear:race and anger:gender.

\para{Parameter Sensitivity}

As expected, Figure \ref{fig:fewshot} shows that with an increase in the proportion of the usage of attribute values from the training data, i.e. the size of the seed set used in the semi-supervised approach, the effectiveness of debiasing improves. Using a percentage higher than that of $10$-$20\%$, however, is likely not a practical situation.
It is observed from the left plot of Figure \ref{fig:epsilonVary1} that too small or too large values of the threshold $\epsilon$, tends to decrease the fairness, the best values of which for both unsupervised and semi-supervised approaches fall within the range of 0.3 to 0.4. The right plot of Figure \ref{fig:epsilonVary1} shows that the optimal value of $K$ for $K$-NN based cluster assignment for few-shot debiasing is $3$.



\section{Related Work}


The study \citep{Bolukbasi:2016} reported that skipgram \cite{Mikolov13} exhibits biases in the neighborhoods of gender neutral words, such as the word `programmer' being associated with a higher proportion of male-specific words. The authors of \citep{Bolukbasi:2016}
proposed a transformation based solution to alleviate such biases.
Other gender neutralizing work include those of
\citep{zhang-etal-2019}, where debiasing of word vectors was extended to contextual word embedding, and
\citep{qian-etal-2019} which proposed a method for estimating a gender-neutral language model.
Making a dataset balanced by a selective removal of data instances can lead to mitigating biases for certain tasks, such as that of author verification \citep{bevendorff-etal-2019}. The study \citep{gonen-goldberg-2019} shows that debiasing the embedding space is particularly not suited to reduce biases from downstream tasks.

Among existing work that explicitly uses annotated attribute values to debias prediction tasks include those of \citep{Zafar:2017} that approached debiasing as a constrained optimization problem. The work in \citep{sen} alleviated the limitation of the feature-based approach of \citep{Zafar:2017} by employing an end-end data-driven multi-objective neural model for debiasing. Blodgett et. al. \cite{BlodgettBDW20} is an excellent survey on the topic.

\section{Conclusion and Future Work}
We proposed a debiasing approach which makes use of a simple, yet effective, clustering-based heuristic to identify, and mitigate biases from predicted outputs.
%
The fact that the approach works either without the presence of any attribute metadata or with its presence in small quantities, indicates the feasibility of deploying unsupervised or semi-supervised debiasing mechanisms at scale. 

A possible extension of this work could be to apply a similar idea for automatic data curation targeted towards mitigating the imbalance in the data, thus yielding a further improvement of a model's fairness.

\bibliographystyle{acl}
\bibliography{cikm} 
\end{document}